\documentclass[final,12pt]{clear2024} %

\usepackage{times}
\usepackage{latexsym}
\usepackage[T1]{fontenc}
\usepackage[utf8]{inputenc}
\usepackage{microtype}
\usepackage{inconsolata}
\usepackage{tikz}
\usepackage{cleveref}
\usepackage{booktabs}
\usepackage{dsfont}
\usepackage{xcolor}
\usepackage{subcaption}     %
\usepackage{enumitem}
\usepackage{makecell}
\usepackage{MnSymbol} %
\usepackage{fnpct} %

\newtheorem{fact}{Fact}[section]

\newtheorem{assumption}[fact]{Assumption}

\definecolor{sbsblue}{HTML}{1f77b4}
\definecolor{sbsorange}{HTML}{ff7f0e}
\definecolor{sbsgreen}{HTML}{2ca02c}
\definecolor{sbsred}{HTML}{d62728}

\usetikzlibrary{cd,arrows,calc,shapes,positioning}
\tikzstyle{obs} = [circle,fill=white,draw=black,inner sep=0pt,minimum size=18pt,font=\fontsize{10}{10}\selectfont,node distance=1,thick]
\tikzstyle{latent} = [obs,dotted]

\newcommand{\nconfounders}[1]{18}

\newcommand{\cc}{\operatorname{CC}}

\newcommand{\atet}{\operatorname{ATET}}

\title[Estimating the Causal Effect of Early ArXiving on Paper Acceptance]{Estimating the Causal Effect of Early ArXiving on Paper Acceptance}
\usepackage{times}

\clearauthor{Yanai Elazar\textsuperscript{1,2}$\thanks{Equal contribution; random ordering. Work done while JZ interned at AI2.}$ \quad
Jiayao Zhang\textsuperscript{3}$\footnotemark[1]$
\quad
David Wadden\textsuperscript{1}$\footnotemark[1]$\quad
{
\bf Bo Zhang\textsuperscript{4} \quad
\bf Noah A.~Smith\textsuperscript{1,2}
}\\
\textsuperscript{1}Allen Institute for Artificial Intelligence \\
\textsuperscript{2}Paul G.~Allen School of Computer Science \& Engineering, University of Washington \\
\textsuperscript{3}University of Pennsylvania \\
\textsuperscript{4}Fred Hutchinson Cancer Center \\
{\tt  yanaiela@gmail.com \quad jiayaozhang@acm.org \quad davidw@allenai.org}\\
}

\begin{document}

\maketitle

\begin{abstract}

What is the effect of releasing a preprint of a paper before it is submitted for peer review?  
No randomized controlled trial has been conducted, so we turn to observational data to answer this question.
We use data from the ICLR conference (2018--2022) and apply methods from causal inference to estimate the effect of arXiving a paper before the reviewing period (\textit{early arXiving}) on its acceptance to the conference.
Adjusting for confounders such as topic, authors, and quality, we may estimate the causal effect.
However, since quality is a challenging construct to estimate,
we use the \textit{negative outcome control} method, using \textit{paper citation count} as a control variable
to debias the quality confounding effect.
Our results suggest that early arXiving may have a small effect on a paper's chances of acceptance. However, this effect (when existing) does not differ significantly across different groups of authors, as grouped by author citation count and institute rank. This suggests that early arXiving does not provide an advantage to any particular group.\footnote{Code and data can be found at: \url{https://github.com/allenai/anonymity-period}.} %

\end{abstract}


\section{Introduction} \label{sec:intro}

In double-blind peer review, the identities of paper authors and reviewers remain hidden. However, due to the rapidly increasing popularity of the online preprint server arXiv\footnote{\url{arxiv.org}} many papers are now available as preprints before they have been submitted to conferences for peer review.

While this process does not directly break double-blind review, the availability of arXiv preprints makes it possible for reviewers to determine the identity of the authors associated with an arXived paper submission. Reviewers, in turn, may be biased based on characteristics of the paper authors such as their citation counts, or their institutions' rankings. Prior studies comparing single- and double-blind review provide mixed evidence, with some pointing to bias in favor of highly-ranked authors or universities \citep{Tomkins2017SingleVD} and others failing to find such a bias \citep{Madden2006ImpactOD}. 
Crucially, no prior work has studied the effect of arXiving (publishing on arXiv) before a submission deadline---which may or not break double-blindness in practice---on paper acceptance. %

\begin{figure}[!t]
    \centering
\begin{center}
\begin{tikzpicture}[node distance=2.2cm,>=latex, scale=1]

\node[obs, fill=sbsblue, fill opacity=0.5, text opacity=1] (arxiv) [] at (0, 0) {$A$};
\node[obs, fill=sbsorange, fill opacity=0.5, text opacity=1] (confounder) [] at (0, -1.5) {$C$};
\node[latent] (latent) [] at (0, 1.5) {$U$};
\node[obs, fill=sbsgreen, fill opacity=0.5, text opacity=1] (nco) [] at (-2.5, 0) {$N$};
\node[obs, fill=sbsred, fill opacity=0.5, text opacity=1] (acceptance) [] at (2.5, 0) {$Y$};

\node[align=left, inner sep=0.5ex, scale=0.6] at (1.7, 1.5) {{\bf Unobserved Confounders:}\\ Creativity, originality};

\node[align=left, inner sep=0.5ex, scale=0.6] at (-0.89, 0) {{\bf Treatment}: \\ Early ArXiv};

\node[align=left, inner sep=0.5ex, scale=0.6] at (-3.2, 0) {{\bf Negative} \\ {\bf Control} \\ {\bf Outcome:}\\ Citation count};

\node[align=left, inner sep=0.5ex, scale=0.6] at (3.5, 0) {{\bf Outcome:} \\ Acceptance};

\node[align=left, inner sep=0.5ex, scale=0.6] at (1.6, -1.6) {{\bf Observed Confounders:}\\ Topic, authors, institute};

\draw[->] (arxiv) -- (acceptance);
\draw[->] (confounder) -- (acceptance);
\draw[->] (latent) -- (acceptance);
\draw[->] (confounder) -- (arxiv);
\draw[->] (latent) -- (arxiv);
\draw[->] (latent) -- (nco);
\draw[->] (confounder) -- (nco);
\path[draw,dashed,-] (acceptance) to [out=150,in=30] (nco);

\end{tikzpicture}
\end{center}
\caption{\textbf{Causal graph of our problem}. %
$A$ and $Y$ are binary treatment and effect variables, respectively: whether a paper was arXived before the review deadline, and whether the paper was accepted. As we cannot measure the unobserved confounders (e.g., quality), we estimate the effect of arXiving using a negative control outcome variable ($N$). Solid edges represent a directed causal effect, while dashed edges represent an association.
}
\label{fig:graph}
\vspace{-5mm}
\end{figure}
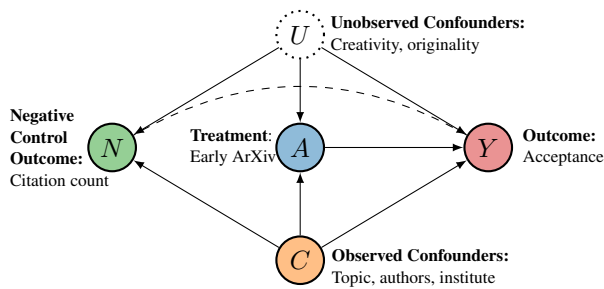

Motivated by these observations, we aim to answer the following questions: \textbf{(RQ1)} Does early arXiving result in different effects on authors based on their citation counts and institutions? \textbf{(RQ2)} What effect (if any) does early arXiving have on paper acceptance? %

To address these questions, we take advantage of a dataset encompassing five years of paper submissions to the ICLR conference. ICLR is unique because is is the only conference which (to our knowledge) releases the acceptance outcomes (accept or reject) of all submitted papers; the availability of these acceptance decisions enables the causal analysis performed in this work.
\footnote{\url{iclr.cc}}
We operationalize the problem as a causal graph (Figure \ref{fig:graph}) where paper acceptance (the \emph{outcome} variable) is determined by three key factors: (1) whether or not the paper was arXived prior to the review deadline (the \emph{treatment} variable), (2) \emph{observed confounders} like paper topic, author citations, and host institution rankings, and (3) \emph{unobserved confounders} like paper quality and potential for impact.

Unobserved confounders present a challenge, since they affect acceptance decisions, but we cannot directly adjust for them since they cannot be straightforwardly measured. To address this challenge, we employ a well-established causal inference technique known as the \emph{negative control outcome} (NCO).\footnote{Causal inference literature also refers to \emph{negative outcome control} (NOC) as the methodology, and NCO as the outcome variable. In this paper we use NCO for both meanings interchangeably for improved readability.} This approach requires identifying an NCO variable which is affected by the same confounders as paper acceptance (our outcome variable), but is \emph{not} affected by early arXiving (our treatment variable). We propose to use a paper's citation count in the years following its initial release as an NCO. Specifically, we binarize citation count to categorize papers as ``highly-cited'' and ``less-cited'', eliminating minor variations in counts. %
We conduct multiple analyses to ensure our findings are robust to the number of years over which we count citations, and the binarization threshold used.

Overall, we find that \textbf{(RQ1)} under standard assumptions, we observe no statistically significant evidence that the effect of early arXiving on acceptance is different for authors with different citation counts or institution ranks, and \textbf{(RQ2)} with additional assumptions, the \emph{effect size} of early arXiving on acceptance is either small (less than $4\%$ in seven out of nine settings), or not statistically significant (in four out of nine settings).

\section{Problem Formulation} \label{sec:formulation}

In this section, we formalize the paper acceptance process using the framework of causal inference.
We aim to characterize the effect of the \textbf{treatment} $A$---a binary variable indicating whether the authors submitted their work to arXiv before the \textbf{reviewing deadline}\footnote{We deliberately choose the reviewing deadline and not the submission deadline to account for papers posted on arXiv between these dates, effectively de-anonymizing the authors.}---on the binary \textbf{outcome} $Y$---whether the paper was accepted. We refer to $A$ as ``early arXiving'' for convenience.

There are a number of factors other than $A$ that may affect whether a paper is accepted. These factors, usually called \textit{confounders} or \textit{covariates}, can be divided into two groups. \textbf{Observed confounders} $C$ are attributes that we can operationalize, explicitly measure, and include as variables in our analysis. These include features like paper topic and writing style,\footnote{Text features were shown to be important predictors for the problem of citation prediction \citep{yogatama2011predicting}.} as well as attributes of the paper authors and their institutions. We list the 18 confounders we consider in Table \ref{tab:confounders-detail}.
\textbf{Unobserved confounders} $U$ are attributes that are challenging or impossible to measure, but which nonetheless may impact the outcome $Y$ as well as the treatment $A$. Possible unobserved confounders in this work include characteristics like creativity, novelty, and potential impact; we refer to these collectively as paper \emph{quality}. Although we cannot measure these characteristics, we present an approach to debias their effect in \Cref{sec:estimation}. 
Finally, our analysis also makes use of paper \emph{citation count}. This variable is used to construct an NCO $N$, which we discuss in depth in \Cref{sec:estimation}. 
$N$ can be associated with the acceptance outcome $Y$, both because a paper's acceptance may increase its citation count, 
and also due to shared confounders such as quality. As such, these variables may be correlated, but this does not affect our estimation discussed in the next section.

\begin{table}[t]
    \centering
    \footnotesize
    \begin{tabular}{lll}
        \toprule
        Variable name                   & Description                                                          & Data type               \\
        \midrule
        \texttt{year}                   & Year in which the paper first became available online.                     & Date                 \\
        \texttt{n\_fig}                 & Number of figures appearing in the paper.                            & Integer                 \\
        \texttt{n\_ref}                 & Number of references (citations) in the paper.                       & Integer                 \\
        \texttt{n\_sec}                 & Number of sections in the paper.                                     & Integer                 \\
        \texttt{log\_text\_length}      & Logarithm of the number of tokens.                                    & Float                   \\
        \texttt{text\_ppl}              & Text perplexity, $2^{-\frac{1}{n} \sum_{i=1}^n \log_2 \mathit{LM}(w_i \mid w_{1:i-1})}$. & Float                   \\
        \texttt{topic\_cluster}         & Topic cluster (20 total) assigned to a paper.                         & Categorical             \\
        \texttt{n\_author}              & Total number of paper authors.                                       & Integer                 \\
        \texttt{n\_author\_female}      & Number of female paper authors.                                      & Integer                 \\
        \texttt{first\_author\_female}  & Indicator whether the first author is female.                        & Binary                  \\
        \texttt{any\_author\_female}    & Indicator whether there are any female authors.                      & Binary                  \\
        \texttt{no\_US\_author}         & Indicator whether there are no US-based authors.                           & Binary                  \\
        \texttt{log\_inst\_rank\_min}   & Log of lowest rank of institution affiliated with any paper author.  & Float                   \\
        \texttt{log\_inst\_rank\_avg}   & Log of average rank of author institutions.                          & Float                   \\
        \texttt{log\_inst\_rank\_max}   & Log of highest rank of institution affiliated with any paper author. & Float                   \\
        \texttt{log\_author\_cite\_min} & Citation count of least-cited author in log scale.                             & Float                   \\
        \texttt{log\_author\_cite\_avg} & Average author citation count in log scale.                                & Float                   \\
        \texttt{log\_author\_cite\_max} & Citation count of most-cited author in log scale.                             & Float                   \\
        \bottomrule
    \end{tabular}
    \caption{\textbf{Observed confounders $C$ included in our analysis}. ``Institute rank'' is computed by counting the total number of accepted papers at ICLR from a given institution, in the two years prior to each submitted paper. ``Topic clusters'' include common AI topics such as ``transfer learning'' and ``language models''. Gender information is provided by the authors at the creation of their OpenReview profiles. See Table \ref{tab:tableone} for the full list of topics, along with additional details and statistics on all confounders.}
    \label{tab:confounders-detail}
\end{table}

\section{Causal Inference: Background} \label{appx:causal_inference} \label{sec:causal:background}

We briefly review the potential-outcomes framework, also known as the Rubin causal model \citep{rubin1974estimating,rubin2005causal} as it applies to our setting. A comprehensive overview is given by \citet{imbens2015causal}; for discussions on causal graphs, see e.g., \citet{pearl1995scm};
for a review on causal inference methods for NLP see e.g., \citet{Feder2021CausalII}; 
for a discussion on using causal inference on textual data, see e.g., \citet{zhang2022some}.

The outcome we study in this work is a binary variable indicating whether a submitted paper is accepted at a conference ($Y$). For a given paper there exist two \emph{potential outcomes}: $Y_{A=1}$ is the acceptance outcome that would occur \emph{if the paper were arXived before the review deadline} --- i.e., \emph{if $A$ were 1}. Similarly, $Y_{A=0}$ is the outcome that would occur \emph{if $A$ were 0}. In practice, we observe exactly one of these two outcomes for a given paper: either the authors arXived it early or they did not. The goal of causal inference is to reason about \emph{counterfactuals}: how would a paper's chances of acceptance have changed if the authors had made the other choice?

The most straightforward way to reason about this counterfactual is with the \emph{average treatment effect} (ATE), defined as:
\begin{equation*} \label{eq:estimand}
    \operatorname{ATE} = \mathbb{E} [Y_{A=1}-Y_{A=0}],
\end{equation*}
which is the treatment effect averaged over all individuals from a population.
In our study, we consider a popular alternative, the \emph{average treatment effect on treated} (ATET):
\begin{equation}
    \atet = \mathbb{E} [Y_{A=1} - Y_{A=0} \mid A = 1],
\end{equation}
which in our case helps to answer \emph{whether early arXiving has created an advantage for the group of submissions that, in fact, arXiv early.}\footnote{In what follows, we simplify notation from $Y_{A=a}$ to $Y_a$ for brevity when there is no potential ambiguity.}

\subsection{Assumptions} \label{app:ci:assumptions}
Our modeling assumptions are described in the causal graph (\Cref{fig:graph}) and we make several standard identification assumptions:
\emph{ignorability}, \emph{positivity}, and \emph{consistency}. In addition, we assume \emph{negative control condition} when using the NCO framework. In what follows we formally define and explain these assumptions.
\begin{assumption}[Identification Assumptions]
\mbox{}
\begin{enumerate}
    \item Ignorability: $\{Y_0, Y_1\} \upmodels A \mid (C, U)$.
    \item Positivity: $0 < \mathbb{P}(A=1 \mid C=c, U=u) < 1$.
    \item Consistency: $Y_a = Y^{\text{obs}}$ if $A^{\text{obs}}=a$.
    \item Negative control: $N_a = N$ for $a=0,1$.
\end{enumerate}
\end{assumption}
Intuitively, \emph{ignorability} assumes the observed and unobserved covariates $(C,U)$ are sufficient to deconfound the effect of the treatment (there are no other unobserved confounders); \emph{positivity} means each unit (``submission'') has a non-zero chance of being early arXived and a non-zero chance of not being early arXived; \emph{consistency} establishes the relationship between potential-outcome $Y_a$ and observed outcome $Y^{\text{obs}}$ under the observed treatment level $A^{\text{obs}}$: the potential-outcome under $A=a$ agrees with the observed outcome when $A=a$.
Finally, we make use of another assumption for dealing with unobserved covariates such as quality;
\emph{negative control condition} dictates that the NCO is not causally affected by the treatment.

\section{Estimating the Causal Effect}\label{sec:estimation}

\paragraph{Causal Estimand}
We write $Y_{A=a}$
as the \emph{potential-outcome} of $Y$
under the treatment $A=a$ for $a=0,1$. Our goal is to estimate the average treatment effect on the treated,

\begin{equation*}
\begin{split}
    \atet = \mathbb{E}[Y_{A=1}-Y_{A=0} \mid A=1].
\end{split}
\end{equation*}
ATET assesses whether early arXiving has created an advantage for the submissions which were arXived early. %
The ATET ranges between $-100\%$ to $100\%$, where larger absolute values indicate a stronger effect of the treatment (either positive or negative), whereas values closer to $0$ indicate weak to no effect.
Our modeling assumptions are detailed in the causal graph (\Cref{fig:graph}). We make several standard identification assumptions (including one for the NCO), discussed in detail in \Cref{app:ci:assumptions}.

\subsection{Effect Estimation}

To estimate the ATET using NCO, we make use of a common technique known as \emph{Difference-in-Difference} 
\citep{lechner2011did,Sofer2016OnNO}.
Additionally, we perform statistical matching \citep{rosenbaum1983central,rosenbaum1989optimal,rubin2005causal} to ensure that our treated and control (i.e., early arXiv vs.~not) groups are comparable. 
All of our subsequent analyses are conducted on the matched sample.

\paragraph{Negative Control Outcome and Difference-in-Difference}
NCO has been studied extensively and used in various domains to debias the effect estimate under the presence of unknown confounding \citep{rosenbaum1989effects,weiss2002specificity,Lipsitch2010NegativeCA,shi2022selective}, where NCO variables are usually motivated from the nature of biological mechanisms \citep{lu2008bio}.
Difference-in-Difference (DiD) is a popular way of estimating ATET in longitudinal studies when the outcome is measured in two time periods $t=0,1$; its history dates back to the nineteenth century \citep{snow1854cholera,card1993minimum,bertrand2004did,lechner2011did}. In this setup four potential-outcomes $Y_{a}(t)$ for $a,t=0,1$ are defined and the ATET can be rewritten as:
\begin{align*}
    \atet =\mathbb{E}[Y_1(1)-Y_0(1)]-\mathbb{E}[Y_1(0)-Y_0(0)].
\end{align*}
A linear outcome model is usually assumed in the DiD literature, which implies the following condition
that relates the measurements over time:
\begin{assumption}[Additive Equi-confounding]
\begin{align*}\mathbb{E}[Y_a(1)-Y_a(0)\mid U, A=a, C]
=
\mathbb{E}[Y_a(1)-Y_a(0)\mid A=a, C],
\end{align*}
for $A=0,1.$
\end{assumption}
Under this assumption, ATET can be estimated through taking the difference (over treatment levels) of the difference (over time periods), hence its name ``difference-in-difference''.
In the case of binary outcome variables, one may assume $Y_a(t)$ follows a logistic model. The ATET can then be estimated
by computing the conditional means (i.e., using the fitted model to predict the probability that $Y_a(t)$ assumes a certain value), and confidence intervals can be estimated through bootstraping.

\citet{Sofer2016OnNO} establish the equivalence between NCO and DiD-based estimators by inspecting their identification assumptions and note that the role of an NCO variable $N$ is equivalent to $Y(0)$ (the outcome at the first time period $t=0$), thus the ATET can
be rewritten, in the case for an NCO $N$, as
\begin{align*}
    \atet =\mathbb{E}[Y_{1}-N_{1}]-\mathbb{E}[Y_0-N_0].
\end{align*}
Given the additive equi-confounding assumption that $Y$ and $N$ are affected by $U$ in a similar way, a debiased ATET can be estimated
by taking the difference (across the treated and control groups) of the difference between $Y$ and $N$. 
Combining statistical matching and DiD renders our analysis more robust to model misspecifications \citep{rubin1979using}.

\paragraph{Matching}
We use statistical matching \citep{rosenbaum1983central,rosenbaum1989optimal,rubin2005causal} to ensure our observed covariates (listed in Table \ref{tab:confounders-detail}) are \emph{balanced} in the treated and matched control groups, thus making the two groups comparable. 
We apply matching to obtain
a matched pair in the control group for each submission in the treated group. 
This ensures each matched pair is comparable in terms of covariates and only differs by the treatment level.
We follow \citet{Chen2022AssociationBA} in using the tripartite matching algorithm  \citep{zhang2021matching}, to ensure (1) numerical variables are matched (with a penalty on their $L_2$ difference); (2) categorical variables \verb|n_author| and \verb|year| are nearly exactly matched; (3) the distribution of \verb|topic_clusters| in the matched group is similar to that in the treated group. 
 After matching, all $1,486$ treated units are matched, and the discrepancy between covariates across treated and matched control groups are significantly reduced, as shown in \Cref{tab:tableone} in the Appendix.

\subsection{Choosing a Negative Control Outcome Variable}

Paper quality is a crucial unobserved confounder for paper acceptance decisions.
We use the NCO framework
\citep{rosenbaum1989effects,Lipsitch2010NegativeCA}
to adjust for the confounding effect of the unobserved confounders $U$.
By identifying a variable $N$ that shares the same observed and unobserved confounders as our outcome $Y$, while being not \emph{causally affected} by the treatment $A$\footnote{$N \upmodels A \mid C, U$} (also entailed by \Cref{fig:graph}), %
we can then attempt to correct the bias due to unobserved confounders. %

In this study, we define an NCO, denoted $N^{(n)}_q$, based on a paper's citation count in the $n$ years following its \textit{initial} release, denoted as $\cc^{(n)}$.\footnote{Note that citation count is \emph{not} to be confused with a measure of paper quality.} \textbf{Importantly}, we retrieve the date of each paper's \textit{first} public appearance---whether on arXiv, in conference proceedings, or some other source (see Appendix \ref{app:paper_citations} for details)\footnote{For instance, if a paper first appeared publicly as an arXiv preprint on May 15, 2018, then $\cc^{(2)}$ for this paper counts the number of citations as of May 15, 2020.}.  Measuring citations in a fixed time window since each paper first became available online---rather than by calendar year or starting from the date of the conference in which the paper was published---eliminates a source of confounding where early-arXived papers would have higher values for $\cc^{{(n)}}$ simply because they were available earlier.

For a given paper, $N^{(n)}_q = 1$ if the paper's citation count $n$ years after its release is above the $q$th quantile of citation counts for all papers in the relevant sample, $\cc^{(n)}_q$. Intuitively, $N^{(n)}_q$ captures whether a given paper was ``highly-cited'' or ``less-cited'' after $n$ years. In \Cref{sec:analysis}, we conduct analyses using values of $\{1, 2, 3\}$ for $n$, and $\{50\%, 75\%, 90\%\}$ for $q$.

\paragraph{Validity of Citation Count as a Negative Control Outcome}

By choosing $N^{(n)}_q$ as our NCO, we are assuming that whether or not a paper happened to be arXived early does not have an appreciable effect on the paper's citation count in a fixed time window following its appearance. We address two possible concerns with this assumption.

First, prior work by \citet{Feldman2018Association} has observed an association between early arXiving and increased citations in the calendar year of the paper's release. These findings do not contradict our NCO assumption, for two reasons: (1) \citet{Feldman2018Association} recognize that their findings are not causal, since they do not account for the presence of unobserved confounders in their analysis, such as the paper's quality. (2) The choice to use citations in the calendar year of publication was necessitated by the lack of availability of exact publication dates in the Semantic Scholar search engine\footnote{\url{semanticscholar.org}} at the time the work was performed. Fortunately, exact publication dates are now available, eliminating the potential confounding effects of measuring citations by calendar year. 

Second, researchers \citep{Feldman2018Association,goldberg2018adversarial} have voiced concerns that arXiving can be used as a technique for ``flag-planting'', whereby authors hope to take credit for an idea (and receive more citations) by posting it on arXiv, while other researchers with the same idea wait for their papers to be reviewed. While this is a plausible concern, we are unaware of any work that has empirically demonstrated a flag-planting effect, and are skeptical that such an effect would be so widespread as to invalidate citation count as an NCO. We welcome an empirical study to quantify the effect of flag-planting in future work. Fortunately, should an even more suitable NCO become available, it can be easily substituted into the causal framework proposed in this work.

\section{The Effect of arXiving on Acceptance} \label{sec:analysis}

We begin by describing the data used for our study (\S \ref{subsec:data}).
Then, we estimate the effect of early arXiving on paper acceptance
while controlling for observed confounders, but \emph{not accounting} for unobserved confounders (\S \ref{subsec:primary}). 
This provides us with a so-called \emph{primary analysis} to assess effect sizes in the matched sample, without relying on any NCO assumptions.
We then stratify the analysis by institution ranking and author citation count to address \textbf{RQ1} (\S \ref{subsec:rq1}).
Finally, by accounting for the unobserved confounders using NCO (\S \ref{sec:estimation}) we seek an answer for \textbf{RQ2} (\S \ref{subsec:rq2}).

\subsection{Dataset}\label{subsec:data}

We use the ICLR  2018--2022 database assembled by \citet{Zhang2022InvestigatingFD}, which includes 10,297 papers. 
We select seven submission-related covariates and eleven author-related covariates, shown in \Cref{tab:confounders-detail}. 
For each of 1,486 early arXived papers, we find a matched non-early arXived paper by matching on all observed confounders. Additional details including
the summary of covariates before and after matching (\Cref{tab:tableone}) are given in Appendix \ref{appx:data}. When using citation counts, recent publications where $\cc^{(n)}$ values are undefined are discarded. 
For example, when using $\cc^{(3)}$, only conferences in or before 2020 ($2023-3=2020$) are kept. 

\subsection{Primary Analysis: Controlling for Observed Confounders}\label{subsec:primary}
\begin{figure*}[t!]
    \centering
    \includegraphics[width=\textwidth]{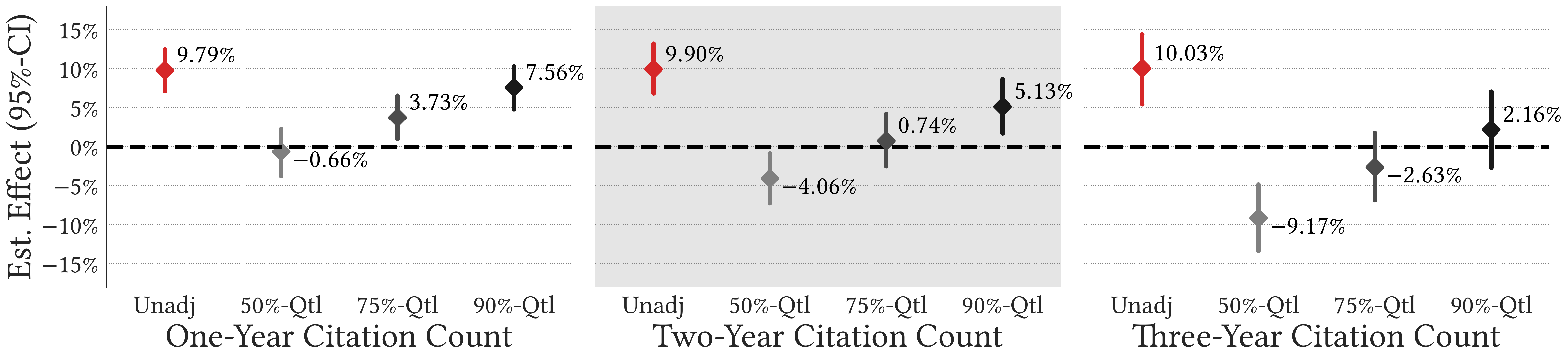}
    \caption{\textbf{Effects of early arXiving (with $95\%$ bootstrap confidence interval) estimated on the matched sample.} 
    ``Unadj'' refers to the estimate without using any NCOs, on the same subset of data that are used by the NCO in the same panel. 
    NCOs are defined to be whether $n$-year citation is greater than the $q$th quantile for $n=1,2,3$ and $q\in\{0.5,0.75,0.9\}$. Estimated effects without debiasing using NCOs are shown in red.
    Note that effects estimated using $N^{(1)}_{0.5}$,
    $N^{(2)}_{0.75}$, $N^{(3)}_{0.9}$, and $N^{(3)}_{0.9}$ are insignificant
    at the $95\%$ level (confidence intervals contain $0$);
    and effects, where significant, are reduced compared with their non-adjusted counterparts (marked in red). This indicates that NCOs help to explain a large part of undebiased effects.
    } \label{fig:fullsmp_fp}
\end{figure*}

We show in \Cref{fig:fullsmp_fp} the estimated effects of early arXiving on the matched sample. Effect estimates from the primary analysis (i.e., controlling only for observed confounders, while not accounting for unobserved confounders) are marked in red (note that the values are different since the sample changes as different $n$-year citation counts are used). The primary analysis on the unmatched groups indicates a significant association between early arXiving and paper acceptance, and the effect size is relatively high: 9.79\%, 9.90\%, and 10.03\% for the samples from the different groups. 
However, in the matched samples from the same groups, early arXived papers are cited almost $2.13$ times more (in a three-year window) than their non early-arXived matches. This suggests that a strong confounder such as paper quality likely exists. We thus attempt to use NCO to study whether the statistically-significant effect of early arXiving effect is attributable to the unobserved confounders.
In what follows we conduct analyses using our data to answer \textbf{RQ1} and \textbf{RQ2}.

\begin{figure*}[t]
    \centering
    \subfigure[{\footnotesize Group by min institution rank, using $N^{(3)}_q$.}]{
        \centering
        \includegraphics[width=0.42\textwidth]{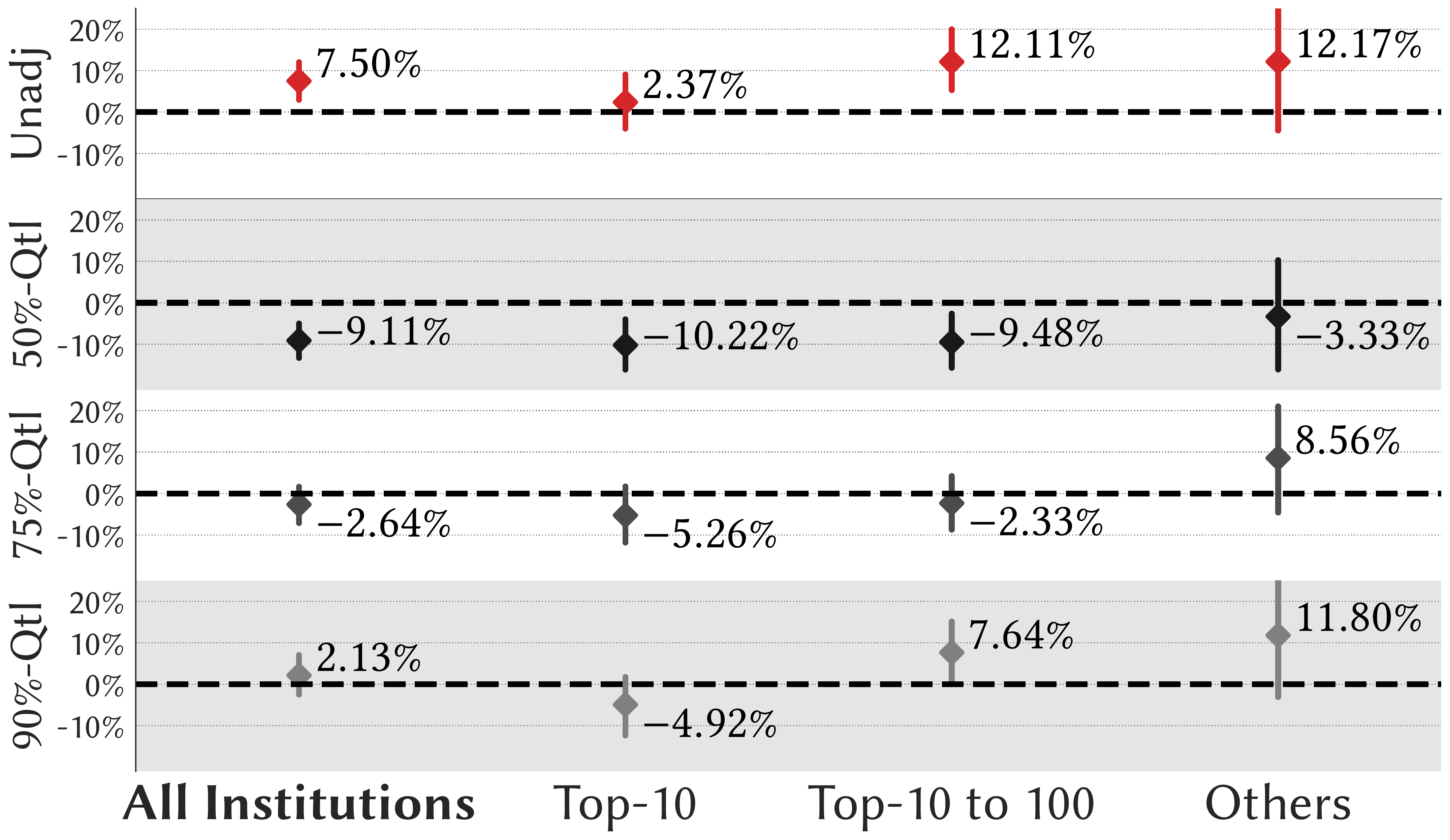}
        \label{fig:n3q1inst}
    }
    \subfigure[{\footnotesize Group by max author citation, using $N^{(3)}_q$.}]{
        \centering
        \includegraphics[width=0.42\textwidth]{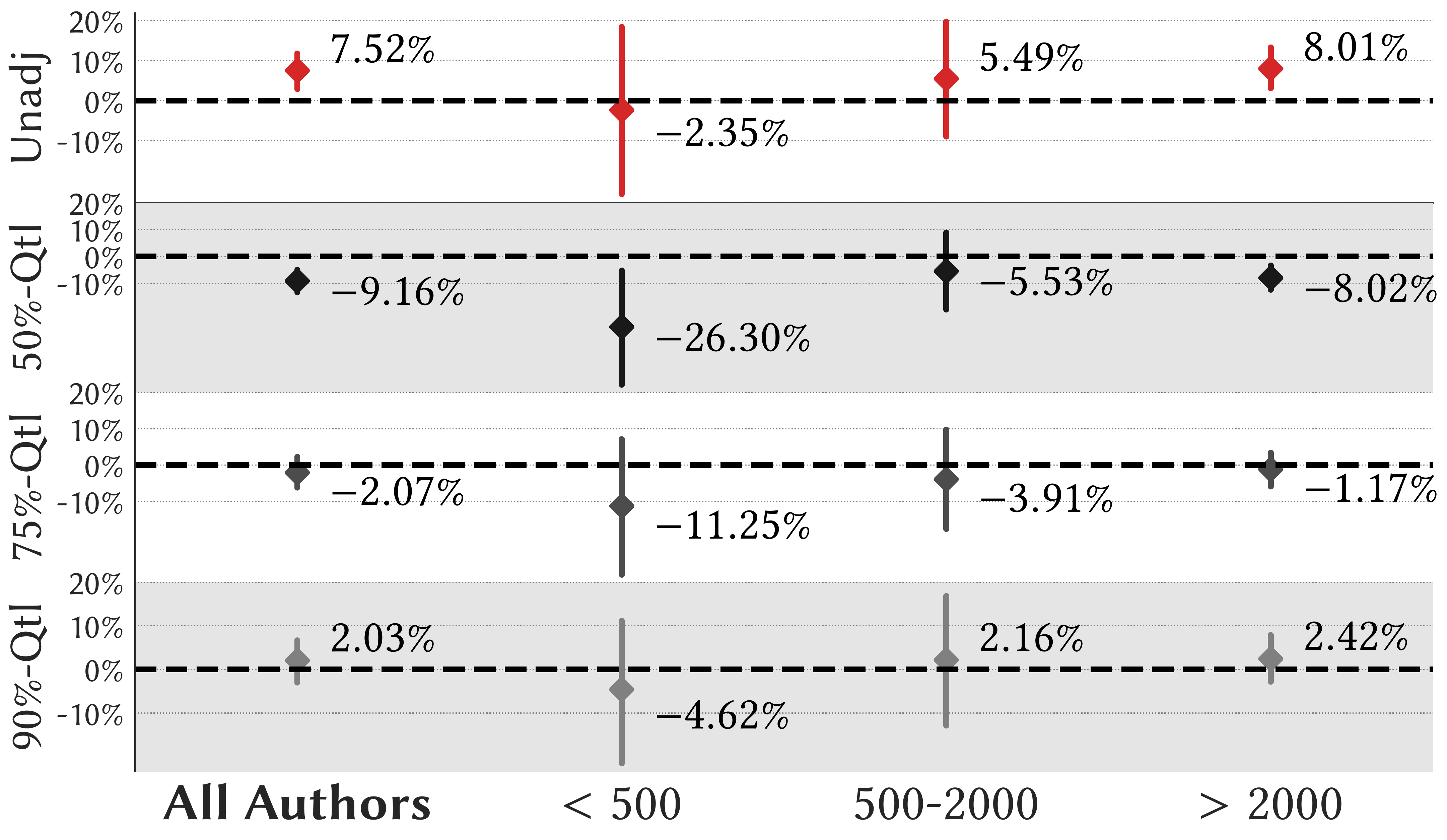}
        \label{fig:n3q9auth}
    }
    \subfigure[{\footnotesize Group by min institution rank, using $N^{(2)}_q$.}]{
        \centering
        \includegraphics[width=0.42\textwidth]{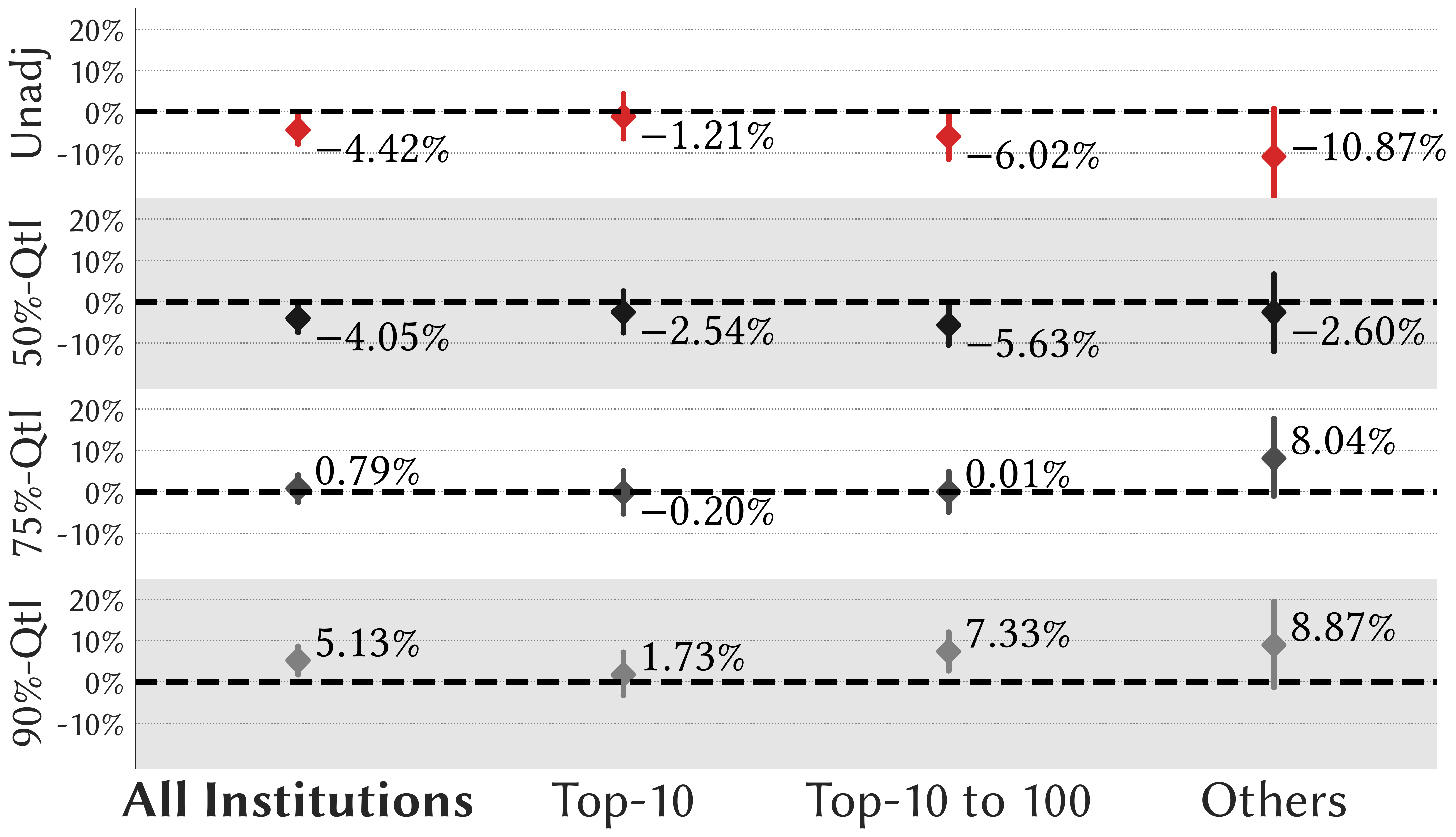}
    } 
    \subfigure[{\footnotesize Group by max author citation, using $N^{(2)}_q$.}]{
        \centering
        \includegraphics[width=0.42\columnwidth]{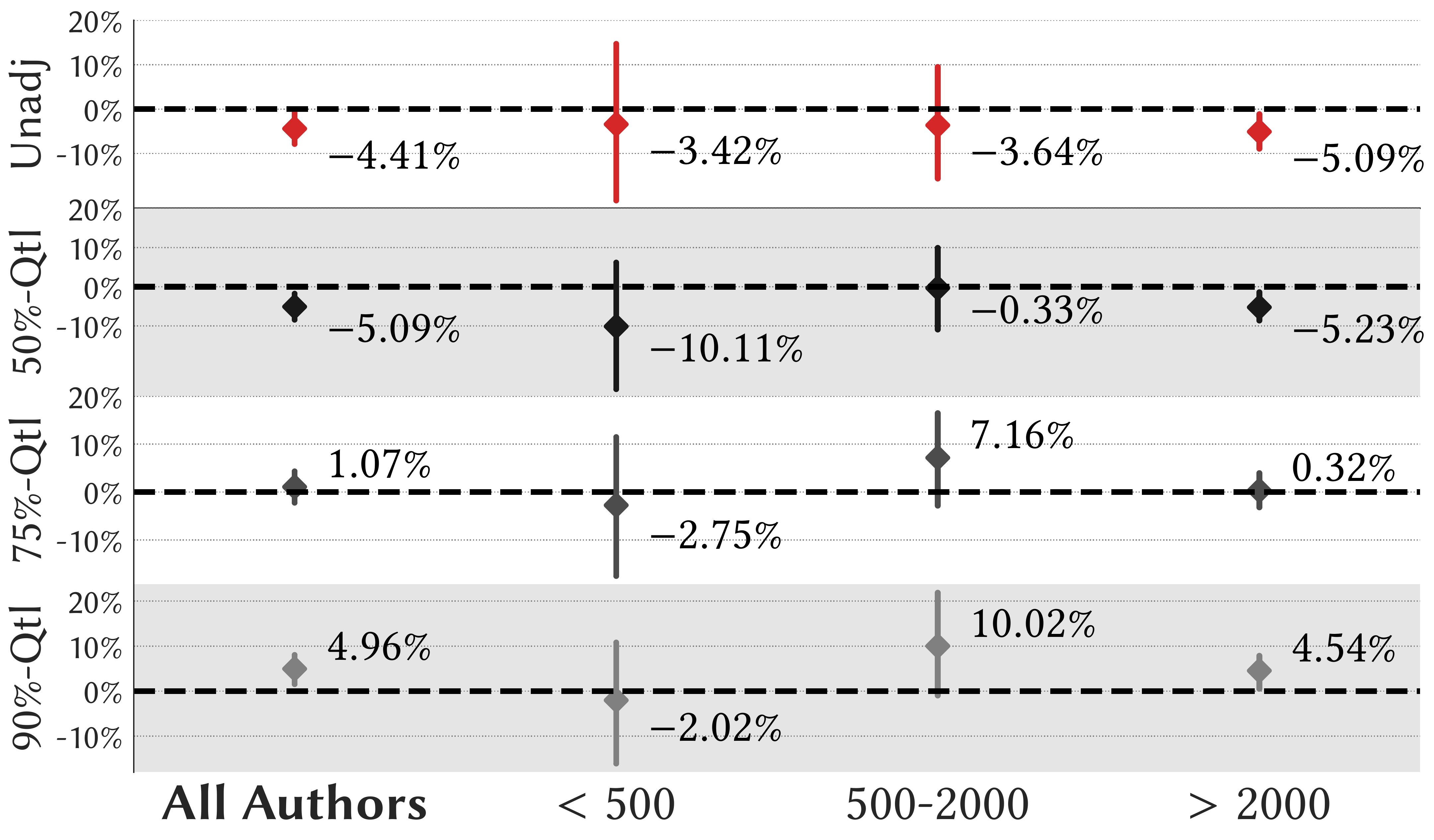}
    }
    \subfigure[{\footnotesize Group by min institution rank, using $N^{(1)}_q$.}]{
        \centering
        \includegraphics[width=0.42\columnwidth]{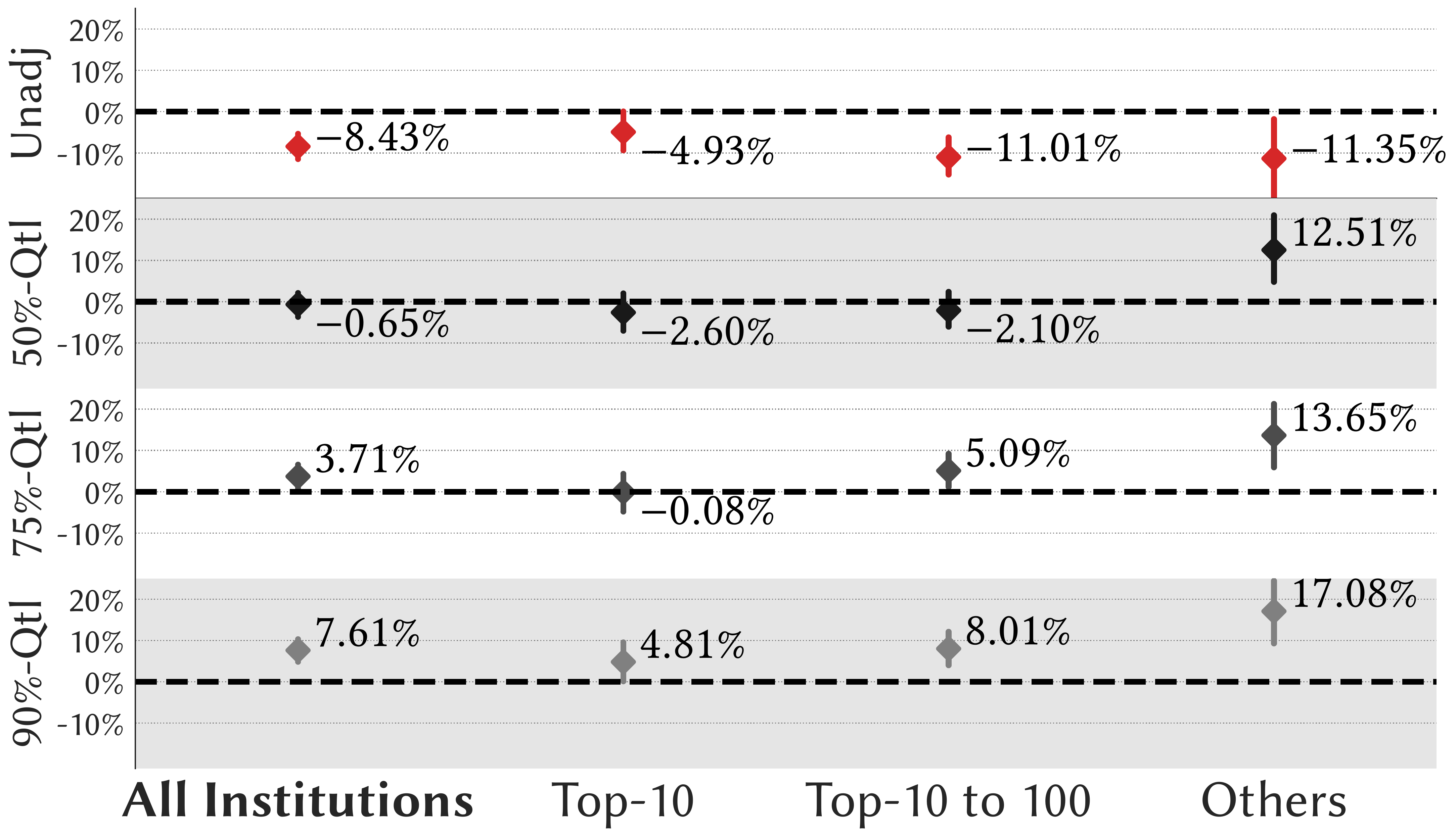}
    }
    \subfigure[{\footnotesize Group by max author citation, using $N^{(1)}_q$.}]{
        \centering
        \includegraphics[width=0.42\columnwidth]{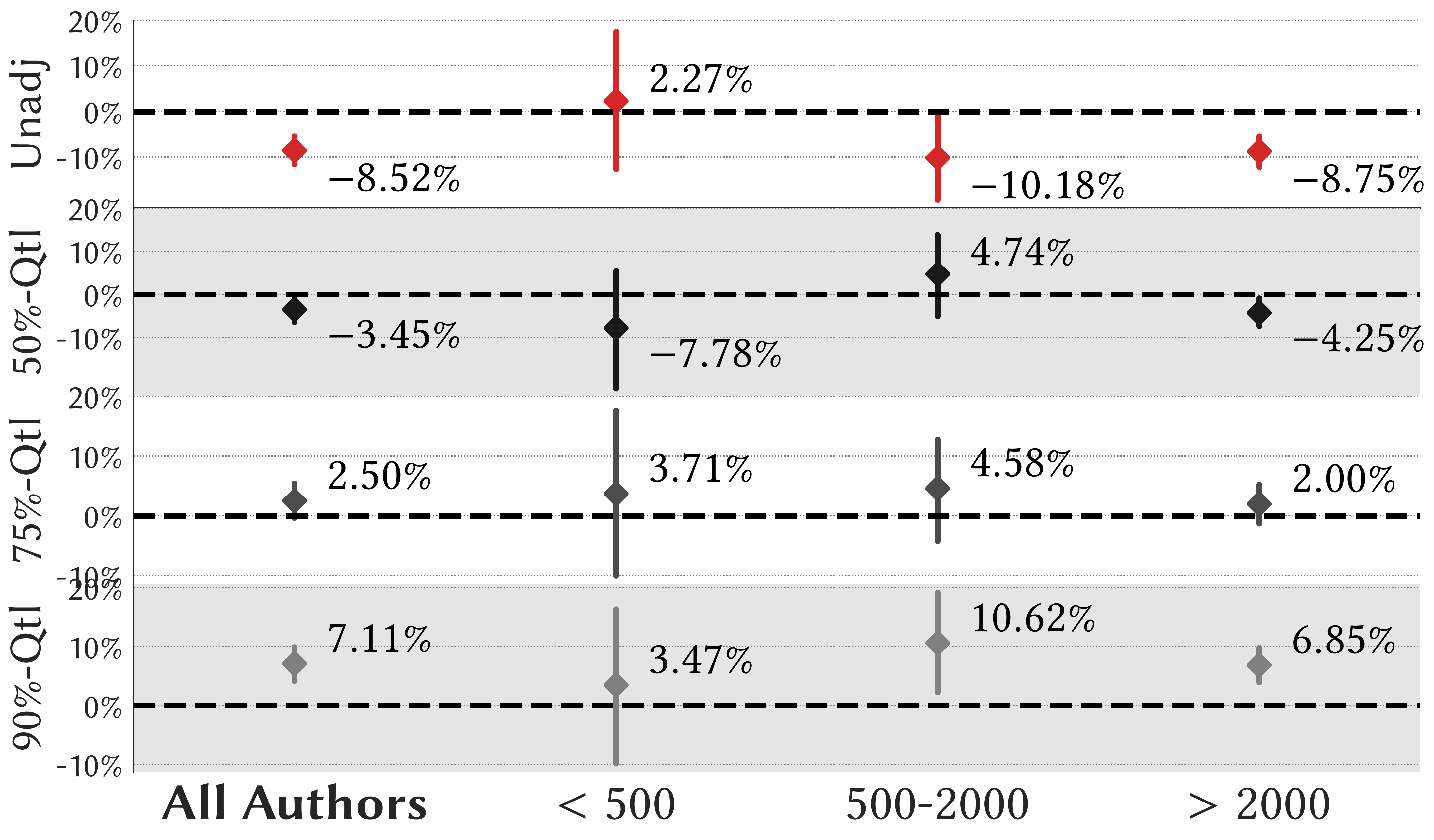}
    }
    \caption{\textbf{Estimated effects in author subgroups on the matched sample.} 
        We use $N^{(n)}$, with three values for $n \in \{1,2,3\}$ as the NCO and estimate the effect %
        across submissions grouped by the minimum author institution and maximum author citation in the submission (``Unadj'' refers to the estimate without using any NCO). Note that
        for NCOs defined using $75\%$ and $90\%$ quantiles, effects are insignificant (confidence intervals contain $0$), and there is no
        evidence that the effects differ across  subgroups (confidence intervals overlap).
        }
    \label{fig:subgrp}
\end{figure*}

\subsection{Analysis on Author Subgroups} \label{subsec:rq1}
One of the concerns regarding the anonymity period in conferences is that arXiving papers may confer an unfair advantage to authors from prestigious institutions, or who are well-known in the community; see \citet{acl2017guide} for ACL Guidelines. %
We acknowledge these aspects have little to do with quality, but are often thought to be biasing factors during the review process \citep{Tomkins2017SingleVD,Snodgrass2006SingleVD}. %
Based on our data and assumptions, \emph{do we have evidence that authors from different groups are treated differently if they early arXiv (\textbf{RQ1})?}

\Cref{fig:subgrp} shows estimated effects %
grouped by institution ranking and author citation count %
(taking min/max respectively in the case of multiple authors).
The lengths of the confidence bands differ, since the set of covariates and sample sizes differ across groups (see \Cref{app:results} for details).

In the primary analysis, without any attempts to adjust for unobserved confounders, we do not find evidence that early arXiving is uniquely beneficial to authors belonging to any specific group, as evidenced by the overlapping confidence intervals for all authors groups (`Unadj', first row of each graph). For instance, when using $N_q^3$ and not adjusting for unobserved confounders, the ATET across all institutions (Figure \ref{fig:n3q1inst}) is 7.5\%; by group, the values are 2.37\%, 12.11\%, and 12.17\% for the top-10, top-11 to 100, and all others, respectively. All confidence intervals overlap, meaning we do not observe statistically significant differences between the different groups. The same can be said when we use NCOs to debias the estimate, e.g., for the 90\% quantile when stratifying between authors citations when using $N_q^3$ (Figure \ref{fig:n3q9auth}), the ATET is 2.03\%, -4.62\%, 2.16\%, and 2.42\% for all authors, authors with less than 500 citations, between 500-2,000 citations, and more than 2,000 citations, respectively. 
These analyses indicate that we do not have evidence that arXiving confers a distinct advantage (in terms of paper acceptance) to any particular author subgroup we consider.

\subsection{NCO Analysis: Accounting for Unobserved Confounders}\label{subsec:rq2}
Finally, we estimate the ATET and $95\%$-bootstrap confidence intervals when the NCO $N^{(n)}_q$ is used in an attempt to adjust for unobserved confounders like paper quality.
The results are shown in \Cref{fig:fullsmp_fp} (the non-red results).
We observe that (1) debiased effects are reduced compared with the estimates when not adjusting for unobserved confounders (marked as ``Unadj'' in the left column), indicating that the simple primary analysis is likely to be confounded and using an NCO helps to deconfound.
(2) Some estimated effects are significant at the $95\%$-level, since their $95\%$-confidence intervals do not contain $0$. This indicates that there are remaining effects that cannot be explained by the NCO we chose. (3) Arguably, the strongest choice of NCO is an $N^{(n)}_q$ with a larger $n$ and higher $q$, which quantifies whether a paper is highly cited over a longer period of time, and thus is less likely to be affected by early arXiving. %
As such, we observe that using $N^{(3)}_{0.75}$ and $N^{(3)}_{0.9}$ result in effect sizes of $-2.63\%$ and $2.16\%$, respectively, which are not significant at the $95\%$ level, indicating the remaining effects are weak.

\section{Conclusion \& Discussion}\label{sec:discussion}

In this work, we investigate whether arXiving a paper before a conference review deadline influences the likelihood that the paper will be accepted at that conference.

We begin with a primary analysis, which assesses the effect of early arXiving on paper acceptance while controlling for observed confounders, but ignoring unobserved confounders such as paper quality. The results of the primary analysis suggest that early arXiving increases a paper's chances of acceptance by roughly $10\%$ on average, a statistically significant effect.

However, when we stratify our analysis across different author cohorts grouped by institution ranking and citation count, we do \emph{not} observe a statistically significant difference in the effect of early arXiving across subgroups. Thus, we do not find evidence that early arXiving provides an unfair advantage to any particular group of researchers, one of the arguments given for the anonymity period of conferences.

Next, in an attempt to account for the effects of unobserved confounders like paper quality, we leverage the NCO framework, and argue for the use of paper citation count after $n$ years as the NCO variable. Using NCO, under a range of reasonable assumptions, we find that early arXiving has a substantially weaker effect on acceptance relative to our primary analysis---increasing acceptance likelihood by less than 4\% in seven out of nine experimental settings (\Cref{fig:fullsmp_fp}). This suggests that papers which were arXived early differ in important but difficult-to-quantify ways from papers which were not, and that our casual inference procedure accounted for some of these differences.

In summation, we find that early arXiving may have a small effect on the likelihood of paper acceptance, under standard assumptions, but that this effect is constant across different researcher subpopulations and does not disadvantage any particular group.
This latter finding is noteworthy because one of the most compelling justifications for conference anonymity periods is that they may prevent particular groups of researchers from gaining an unfair advantage by releasing early preprints of their work. Our findings call this justification into question. Therefore, we would advocate for a randomized controlled trial to examine the effects of early arXiving in a more controlled setting; if such a trial recapitulates the results found here, it would suggest that anonymity periods may not be necessary to preserve fairness across different researcher populations.

\section*{Epilogue}
On January 12, 2024, the day we received the acceptance notification of this paper at CLeaR 2024, the Association for Computational Linguistics decided to remove the anonymity period policy effective immediately from all of its venues.\footnote{\url{https://aclweb.org/adminwiki/index.php/ACL_Anonymity_Policy}.} This paper was greatly influenced by the ACL policy and our curiosity about whether the policy's original goals and motivations were held in practice. We are unlikely to know whether this paper contributed to this policy lift from ACL (we made the paper first available on June 24, 2023). As with many real-world causal questions, assessing the true causal effect is hard.

\section*{Ethics Statement}
Part of the decision for starting the anonymity period was motivated by the negative biases that arXiving might have on the double-blind review process \citep{acl2017guide}. %
Our results suggest that the indirect violation of this process through arXiving does not affect different groups (stratified by instutute or citations) differently. Together with additional assumptions, we observe that the effect (when existing) is small (in seven out of nine settings it is less than $4\%$).
As such, our results should not be taken as a ground truth, but as some initial evidence that anonymity periods (such as the ACL anonymity period) may not achieve one of its main goals.

\section*{Limitations}
This work has a number of limitations. First, we analyze decisions from a single machine learning conference where acceptance decisions are readily available, and thus our results may not generalize to other venues or research communities. Second, the validity of the modeling techniques used in this work rests on assumptions that we believe are plausible, and make the best use of the available data, but which are certainly up for debate.
We attempted to control for as many observed covariates as we could, and observed the effects estimated using NCOs drops compared with the effects estimated without any NCOs. 
Although this suggests that NCOs help to deconfound the effects, the NCO assumption that a long-term (e.g., $3$-year) citation count is not causally affected by early arXiving is uncheckable, and the conclusions are contigent on this assumption.
We welcome future works to estimate such effects with other assumptions.
Ideally, one would conduct a randomized controlled trial (or a randomized encouragement trial) in order to obtain an unbiased estimate of early arXiving effects. We note that such an experiment is not trivial to conduct, as it requires a large enough sample group, consent to participate in an experiment that would randomly assign them to a group and require them to either (1) keep their publication anonymous until the decisions, or (2) arXiv their papers before the reviews.

\acks{We would like to thank the anonymous reviewers, Ashish Sabharwal, and Tim Althoff, as well as the AllenNLP and Semantic Scholar teams for feedback on drafts of this paper.}

\bibliography{bib/anthology,bib/custom,bib/causal}

\setcounter{section}{1}\setcounter{equation}{0}
\numberwithin{equation}{section}
\counterwithin{figure}{section}
\counterwithin{table}{section}
\counterwithin{fact}{section}

\clearpage
\appendix

\section{Data Preparation} \label{appx:data}
\begin{table*}[t]
    \centering
    \resizebox{\linewidth}{!}{%
\begin{tabular}{lllllll}
\toprule
    \textbf{Covariate}                            &      & \shortstack{\bf Early arXived Papers\\(Treated Group, $n=1486$)} & \shortstack{\bf Non-Early arXived Papers\\(Unmatched Control, $n=7493$)} & \shortstack{\textbf{Matched Comparison Group}\\($n=1486$)} & \shortstack{\bf SMD\\(Before Matching)} & \shortstack{\bf SMD\\(After Matching)} \\
\midrule

\texttt{year} & &    &   &                             &                 0.746 &                0.002 \\
\quad \texttt{2018} &&                              20 (1.3) &                             710 (9.5) &                    20 (1.3) &                       &                      \\
\quad \texttt{2019} &&                              48 (3.2) &                           1146 (15.3) &                    48 (3.2) &                       &                      \\
\quad \texttt{2020} & &                           502 (33.8) &                           1422 (19.0) &                  502 (33.8) &                       &                      \\
\quad \texttt{2021} &&                            503 (33.8) &                           1876 (25.0) &                  502 (33.8) &                       &                      \\
\quad \texttt{2022} &&                            413 (27.8) &                           1967 (26.3) &                  414 (27.9) &                       &                      \\
\midrule
\texttt{n\_fig} &      &                            14.3 (7.5) &                            12.6 (7.3) &                  14.0 (8.2) &                -0.221 &                0.035 \\
\texttt{n\_ref} &      &                           44.7 (16.7) &                           40.9 (16.9) &                 44.4 (18.6) &                -0.224 &                0.017 \\
\texttt{n\_sec} &      &                            20.8 (7.8) &                            19.2 (7.1) &                  20.9 (7.9) &                -0.213 &               -0.018 \\
\texttt{log\_text\_length} &      &                             4.0 (0.2) &                             4.0 (0.2) &                   4.0 (0.2) &                -0.355 &               -0.002 \\
\texttt{text\_ppl} &      &                             0.8 (0.0) &                             0.8 (0.0) &                   0.8 (0.0) &                 0.172 &               -0.015 \\
\midrule
\texttt{topic\_cluster} &&&&&                 0.188 &               <0.001 \\
\quad 00 \quad \texttt{RL/DL/Robustness} &&                              77 (5.2) &                             325 (4.3) &                    77 (5.2) &                 & \\
\quad 01 \quad \texttt{RL/DL/CV} &&                              66 (4.4) &                             392 (5.2) &                    66 (4.4) &                       &                      \\
\quad 02 \quad \texttt{DL/Generative Models/CNN} &&                              46 (3.1) &                             223 (3.0) &                    46 (3.1) &                       &                      \\
\quad 03 \quad \texttt{DL/RNN/GNN} &&                              55 (3.7) &                             340 (4.5) &                    55 (3.7) &                       &                      \\
\quad 04 \quad \texttt{DL/Optimization/Generalization} &&                             140 (9.4) &                             495 (6.6) &                   140 (9.4) &                       &                      \\
\quad 05 \quad \texttt{DL/Robustness/Adversarial Examples} &&                             106 (7.1) &                             458 (6.1) &                   106 (7.1) &                       &                      \\
\quad 06 \quad \texttt{DL/RNN/GNN} &&                              87 (5.9) &                             363 (4.8) &                    87 (5.9) &                       &                      \\
\quad 07 \quad \texttt{RL/Multi-Agent RL/DL} &&                             129 (8.7) &                            752 (10.0) &                   129 (8.7) &                       &                      \\
\quad 08 \quad \texttt{Federated Learning/DL} &&                             62 (4.2) &                             271 (3.6) &                    62 (4.2) &                       &                      \\
\quad 09 \quad \texttt{Generative Models/VAE/GAN} &&                              74 (5.0) &                             465 (6.2) &                    74 (5.0) &                       &                      \\
\quad 10 \quad \texttt{DL/NLP/Transformer/LM} &&                             108 (7.3) &                             543 (7.2) &                   108 (7.3) &                       &                      \\
\quad 11 \quad \texttt{GNN/GCNN/Representation Learning} &&                              94 (6.3) &                             387 (5.2) &                    94 (6.3) &                       &                      \\
\quad 12 \quad \texttt{DL/Self-Supervised Learning/Meta-Learning} &&                              87 (5.9) &                             455 (6.1) &                    87 (5.9) &                       &                      \\
\quad 13 \quad \texttt{RL/GNN/Transformer} &&                              49 (3.3) &                             289 (3.9) &                    49 (3.3) &                       &                      \\
\quad 14 \quad \texttt{DL/Model Compression/Neural Architecture Search} &&                              48 (3.2) &                             332 (4.4) &                    48 (3.2) &                       &                      \\
\quad 15 \quad \texttt{DL/Representation Learning/Transfer Learning} &&                              44 (3.0) &                             274 (3.7) &                    44 (3.0) &                       &                      \\
\quad 16 \quad \texttt{DL/Representation Learning/Word Embeddings/NLP} &&                              41 (2.8) &                             251 (3.3) &                    41 (2.8) &                       &                      \\
\quad 17 \quad \texttt{DL/Neural Architecture Search/Optimization} &&                              50 (3.4) &                             296 (4.0) &                    50 (3.4) &                       &                      \\
\quad 18 \quad \texttt{DL/RL/Representation Learning/Generalization} &&                              57 (3.8) &                             299 (4.0) &                    57 (3.8) &                       &                      \\
\quad 19 \quad \texttt{DL/Interpretability/Uncertainty Estimation} &&                              66 (4.4) &                             283 (3.8) &                    66 (4.4) &                       &                      \\
\midrule
\texttt{n\_author} &      &                             4.2 (1.9) &                             4.3 (2.0) &                   4.1 (1.9) &                 0.096 &                0.002 \\
\texttt{n\_author\_female} &      &                             0.3 (0.6) &                             0.4 (0.6) &                   0.3 (0.6) &                 0.087 &                0.009 \\
\texttt{first\_author\_female} & && & &                 0.106 &                0.006 \\
\quad \texttt{True} &&                              87 (5.9) &                             643 (8.6) &                    89 (6.0) &                       &                      \\
\quad \texttt{False} & &                          1399 (94.1) &                           6850 (91.4) &                 1397 (94.0) &                 &               \\
\texttt{any\_author\_female} &  &    & & &                 0.085 &                0.022 \\
\quad \texttt{True} &&                            375 (25.2) &                           2173 (29.0) &                  361 (24.3) &                       &                      \\
\quad \texttt{False} &&                           1111 (74.8) &                           5320 (71.0) &                 1125 (75.7) &                 & \\
\texttt{no\_US\_author} &  & & & &                 0.056 &                0.006 \\
\quad \texttt{True} &&                            464 (31.2) &                           2537 (33.9) &                  468 (31.5) &                       &                      \\
 \quad \texttt{False} &&                           1022 (68.8) &                           4956 (66.1) &                 1018 (68.5) &&\\
 \midrule
\texttt{log\_ins\_rank\_min} &      &                             1.1 (0.7) &                             1.1 (0.7) &                   1.1 (0.7) &                 0.042 &               -0.001 \\
\texttt{log\_ins\_rank\_avg} &      &                             1.6 (0.6) &                             1.5 (0.6) &                   1.6 (0.6) &                -0.041 &               -0.000 \\
\texttt{log\_ins\_rank\_max} &      &                             1.8 (0.6) &                             1.7 (0.6) &                   1.8 (0.6) &                -0.064 &               -0.001 \\
\texttt{log\_author\_cite\_min} &      &                             2.3 (0.8) &                             2.3 (0.8) &                   2.4 (0.8) &                 0.008 &               -0.021 \\
\texttt{log\_author\_cite\_avg} &      &                             3.6 (0.6) &                             3.6 (0.7) &                   3.6 (0.6) &                -0.067 &               <0.001 \\
\texttt{log\_author\_cite\_max} &      &                             4.0 (0.7) &                             3.9 (0.8) &                   3.9 (0.7) &                -0.083 &                0.012 \\
\bottomrule
\end{tabular}
}
    \caption{\textbf{Summary of covariates before and after matching.} We show the mean and standard deviation (in parentheses) for each covariate, and the standardized mean difference (SMD)
    between the treated group and unmatched and matched comparison groups. Note that after matching the SMD reduces significantly for all covariates and a fine-balance
    has been achieved in the \texttt{topic\_cluster}.
    In \texttt{topic\_cluster} we also
    display top keywords associated with each cluster. Abbreviations: \texttt{RL = Reinforcement Learning}, \texttt{DL = Deep Learning}, \texttt{CV = Computer Vision},
    \texttt{[GC]NN = [Graph Convolutional] Neural Nets}, \texttt{VAE = Variational Autoencoder}, \texttt{GAN = Generative Adversarial Network}.
    }
    \label{tab:tableone}
\end{table*}
We build on data collected and processed by \citet{Zhang2022InvestigatingFD} using the ICLR data between the years 2018--2022.\footnote{\url{cogcomp.github.io/iclr_database}} The original dataset includes 10,297 submissions between 2017--2022, consisting of all the submitted papers to ICLR between these years, not including retracted papers.  Submissions in the year
of 2017 are not used in the study as the treated group contains too few samples.

\subsection{Confounders}

We consider \nconfounders{} confounders, summarized in \Cref{tab:tableone}. We now describe the processing of these variables, divided into two high level categories: \textit{submission metadata} and \textit{submission content}.

\paragraph{Submission Metadata}
We use the following metadata information: submission year, number of authors, number of female-identified authors,\footnote{The gender is based on author's OpenReview profile. We construct the indicator variable of whether an author is self reported to be female to be used in our analysis.
} whether the first author is identifying as female, and whether there is a non-US base author.
In addition, we use some features that account for the authors institute, and citation count: the highest, lowest, and average institution ranking,\footnote{Institute ranking is determined by counting the total number of papers accepted to ICLR in previous years.} and the highest, lowest, and average citation counts of the authors.\footnote{Author citation counts are obtained from the Google Scholar API \citep{cholewiak2021scholarly}, as of Feburary, 2022.}

\paragraph{Submission Content}
Using the information extracted from the papers, we take into account several content-related potential confounders. We consider the number of figures, references and sections of a paper
extracted using Grobid\footnote{\url{github.com/kermitt2/grobid}} on the fulltext. We also use the length of the paper (in log scale) measured using Longformer tokenizer \citep{Beltagy2020Longformer}, and a measure of paraphrase fluency\footnote{\url{github.com/PrithivirajDamodaran/Parrot_Paraphraser}} averaged over the full document (calculated by averaging over paraphrased token likelihoods from the pre-trained RoBERTa model \citep{roberta}, normalized to $(0,1)$, where higher values indicate the utterances are more likely, For simplicity, we refer to this as \texttt{text\_perplexity}).
Finally, we consider 20 topic clusters, computed by spectral clustering \citep{ng2001sc} using the Sepcter embedding \citep{specter2020cohan} of abstracts in \texttt{scikit-learn} \citep{scikit-learn}.

\subsection{Paper Citations} \label{app:paper_citations}

\begin{table}[t!]
    \footnotesize
    \centering
    \begin{tabular}{lrr}
       \toprule  
       & Accepted & Rejected \\
       \midrule
       \# submissions &  3,678 & 6,619  \\
       Has S2 ID & 3,636 / 3,678 (99\%)  & 5,336 / 6,619 (81\%) \\
       Has date & 3,209 / 3,636 (88\%) & 4,879 / 5,336 (91\%) \\
       \bottomrule
    \end{tabular}
    \caption{Statistics from the Semantic Scholar (S2) data processing.
    We present the total number of considered submissions (\textit{\# submissions}), the fraction of submissions we successfully matched to an S2 document (\textit{Has S2 ID}), and the fraction of submissions with available publication date from the S2 matched submissions (\textit{Has date}).
    }
    \label{tab:s2_matching}
\end{table}

In \Cref{sec:estimation}, the citation count $\cc^{(n)}$ is described as the number of times that a given paper was cited in the $n$-year window after it was first made public. As a concrete example, for a paper that first appeared online as an arXiv preprint on May 15, 2018, $\cc^{(2)}$ counts the number of citations received by this paper as of May 15, 2020. For a paper that was never arXived and first appeared online as part of conference proceedings published on October 20, 2018, $\cc^{(2)}$ counts the number of citations received by this paper as of October 20, 2020. This allows for an ``apples-to-apples'' comparison of papers' citation counts; by contrast, comparing citation counts by (for instance) calendar year would inflate the citation counts of papers preprinted on arXiv prior to their appearance in conference proceedings, relative to papers that did not first appear as preprints.

We describe our process of obtaining $\cc^{(n)}$ using the Semantic Scholar Academic Graph \citep{Kinney2023TheSS}.

\paragraph{Matching ICLR papers to Semantic Scholar}

First, we use the Semantic Scholar (S2) API to match each submitted ICLR paper to a unique document in S2; this is done using a fuzzy match on paper title. The S2 pipeline includes a canonicalization step where multiple instances of the same paper from different sources are merged into a single entry; for instance, a paper that was released first on arXiv and then included in the ICLR conference proceedings is represented as a single paper in S2, and its citations are merged (i.e. citations of both the arXiv preprint and the conference article are included in the canonical citation count).

Next, we query the S2 Academic Graph API for metadata on each matched paper, including its publication date. For S2, the publication date is not the day of the conference at which the paper was presented, but rather the \emph{date that the paper first became available through any of the data sources S2 ingests} (including arXiv, DBLP, the ACL anthology, etc.). Effectively, it is the first day when the paper became visible online.

\Cref{tab:s2_matching} shows the results of these data processing steps. We were able to match nearly all accepted ICLR papers to a paper in S2. For rejected papers, we matched 81\%. The remaining 19\% of rejected papers were likely never posted to arXiv or published elsewhere. In our analysis, we assign these papers 0 citations since they were never published and hence never cited. Of the papers where we found a matching paper in S2, roughly 90\% of both accepted and rejected papers had a publication date available in their metadata; the 10\% with no publication dates were removed. Publication dates were missing in 6\% of the early arXived papers, and 11\% of the rest.

\paragraph{Counting Paper Citations}

For each paper $P$ with an available publication date, we queried the S2 Academic Graph API for all papers citing $P$, and submitted a followup query requesting the publication dates for all citing papers. Roughly 10\% of citing papers did not have publication dates available; these were discarded. With the publication date of $P$ and all its citing papers in hand, we compute $\cc^{(n)}$ by counting the number of citing papers which were published (i.e. became available online) within $n$ years of the publication of $P$.

\section{Setup Details} 
\label{app:results}
\begin{table}[t]
    \centering
    \footnotesize
    \resizebox{0.75\columnwidth}{!}{%
    \begin{tabular}{lccc}
        \toprule
        \textbf{NCO} &  $\cc^{(1)}$ &$\cc^{(2)}$ & $\cc^{(3)}$ \\
        \midrule
        
        \textbf{Conference Years} &  $2018-2022$ & $2018-2021$  & $2018-2020$ \\
        \textbf{Matched Pairs} & $1,486$ & $1,073$ & $570$\\
        \bottomrule
    \end{tabular}
    }
    \caption{\textbf{Number of Matched Pairs.} 
    The covered years of the ICLR conference, together with the matched pairs found in the data for each negative control outcome variable ($CC^{(n)}$) decision (the number of citations after $n\in \{1,2,3\}$ years).
    }
    \label{tab:matchedpairs}
\end{table}
\begin{table}[t]
    \centering
    \footnotesize
    \resizebox{0.75\columnwidth}{!}{%
    \begin{tabular}{llrrrrr}
\toprule
 \textbf{Year}     &       \textbf{\shortstack{Early arXiving\\($A$) }} &  \textbf{\shortstack{\# Accept\\(Total Cnt)}} &  \textbf{\shortstack{\# Reject\\(Total Cnt)}} &   \textbf{\shortstack{$\cc^{(1)}$\\(Average)}} &    \textbf{\shortstack{$\cc^{(2)}$\\(Average)}} &   \textbf{\shortstack{$\cc^{(3)}$\\(Average)}} \\
\midrule
2018 & $A=0$ &   10 &   10 &   9.85 &   27.45 &   47.40 \\
     & $A=1$ &   19 &    1 &  30.10 &  112.60 &  249.95 \\
2019 & $A=0$ &   24 &   24 &  14.52 &   38.70 &   63.52 \\
     & $A=1$ &   27 &   21 &  14.13 &   39.04 &   69.79 \\
2020 & $A=0$ &  190 &  312 &   8.32 &   20.53 &   34.03 \\
     & $A=1$ &  235 &  267 &  11.86 &   34.50 &   58.14 \\
2021 & $A=0$ &  182 &  320 &   8.99 &   25.15 &   - \\
     & $A=1$ &  232 &  271 &   8.66 &   24.33 &   - \\
2022 & $A=0$ &  198 &  216 &   8.76 &   - &    - \\
     & $A=1$ &  236 &  177 &   7.81 &   - &    -  \\
\bottomrule
\end{tabular}
    }
    \caption{\textbf{Summary of paper decision and $n$-year citation counts (average) in the matched sample.} 
    Although the number of accepted papers are 
    larger in the early arXived groups, long-term citation counts are also higher. This suggests that it is likely an unobserved confounder (paper quality) that affect both the paper acceptance and the long-term citation count.
    }
    \label{tab:matchedYN}
\end{table}
\begin{table}[t]
    \centering
    \footnotesize
    \resizebox{0.75\columnwidth}{!}{%
\begin{tabular}{llrrrrrr}
\toprule
            \multicolumn{2}{l}{\bf Author Institution} & \multicolumn{2}{c}{Top-$10$} & \multicolumn{2}{c}{Top-$10$ to $100$} & \multicolumn{2}{c}{Others} \\
            \multicolumn{2}{l}{\bf Early arXiving ($A$)} &    $A=0$ & $A=1$ &             $A=0$ & $A=1$ &  $A=0$ & $A=1$ \\
$\cc^{(n)}$ & Qtl &          &       &                   &       &        &       \\
\midrule
$\cc^{(1)}$ & $50\%$ ($4$) &       36 &    46 &               320 &   398 &    214 &   322 \\
            & $75\%$ ($11$) &       14 &    16 &               171 &   225 &     86 &   139 \\
            & $90\%$ ($23$) &        5 &     4 &                77 &   102 &     26 &    42 \\
$\cc^{(2)}$ & $50\%$ ($10$) &       26 &    39 &               247 &   304 &    141 &   232 \\
            & $75\%$ ($29$) &       13 &    16 &               133 &   172 &     54 &   108 \\
            & $90\%$ ($65$) &        3 &     6 &                52 &    85 &     16 &    34 \\
$\cc^{(3)}$ & $50\%$ ($11$) &       20 &    26 &               168 &   205 &     91 &   157 \\
            & $75\%$ ($41$) &        8 &    11 &                95 &   118 &     32 &    75 \\
            & $90\%$ ($103$) &        2 &     4 &                34 &    59 &      9 &    28 \\
\bottomrule
\end{tabular}
        }
    \caption{\textbf{Sample sizes in each subgroup when stratified by author institution ranking}.
    }
    \label{tab:smp:inst}
\end{table}

\begin{table}[t]
    \centering
    \footnotesize
    \resizebox{0.75\columnwidth}{!}{%
\begin{tabular}{llrrrrrr}
\toprule
            \multicolumn{2}{l}{\bf Author Citation} & \multicolumn{2}{c}{$<$500} & \multicolumn{2}{c}{500-2000} & \multicolumn{2}{c}{$>$2000} \\
            \multicolumn{2}{l}{\bf Early arXiving ($A$)} &  $A=0$ & $A=1$ &    $A=0$ & $A=1$ &   $A=0$ & $A=1$ \\
$\cc^{(n)}$ & Qtl &        &       &          &       &         &       \\
\midrule
$\cc^{(1)}$ & $50\%$ ($4$) &     16 &    25 &       54 &    64 &     500 &   677 \\
            & $75\%$ ($11$) &     10 &     9 &       15 &    25 &     246 &   346 \\
            & $90\%$ ($23$) &      2 &     2 &        7 &     7 &      99 &   139 \\
$\cc^{(2)}$ & $50\%$ ($10$) &     15 &    15 &       40 &    52 &     359 &   508 \\
            & $75\%$ ($29$) &      9 &     7 &       15 &    21 &     176 &   268 \\
            & $90\%$ ($65$) &      2 &     2 &        4 &     6 &      65 &   117 \\
$\cc^{(3)}$ & $50\%$ ($11$) &     15 &    16 &       27 &    32 &     237 &   340 \\
            & $75\%$ ($41$) &      9 &     8 &        9 &    16 &     117 &   180 \\
            & $90\%$ ($103$) &      2 &     2 &        2 &     6 &      41 &    83 \\
\bottomrule
\end{tabular}
}
    \caption{\textbf{Sample sizes in each subgroup when stratified by author citation count}.
    }
    \label{tab:smp:auth}
\end{table}

\paragraph{Matched Pairs}
We perform tripartite matching \citep{zhang2021matching} under the same setup
as in \citet{Chen2022AssociationBA}
such that the (1) discrepancy between numerical covariates is minimal; (2) the categorical variables \verb|n_author| and \verb|year| are nearly exactly matched; (3) the distribution of \verb|topic_cluster| in treated and matched groups are similar (also known as \emph{fine-balance}.
Recall that when performing analysis using NCOs ($\cc^{(n)}$), only a subset of data with conference year in or before the year $(2023-n)$ is used. We tabulate in \Cref{tab:matchedpairs} the sample size (as the number of matched pairs) for each subset.

\paragraph{Motivation for Citation Counts}
As shown in \Cref{tab:matchedYN}, although early arXived papers have higher rate of acceptance in the matched sample, long-term citation counts are also higher. This indicates that it is likely that an unobserved confounder (e.g., paper quality) affects both of them.

\paragraph{Stratification}
When we perform stratified analysis, the same strata may contain a different number of samples depending on the citation counts we use.
The breakdown of sample sizes is tabulated in \Cref{tab:smp:inst} for stratification under author institution ranking, and in \Cref{tab:smp:auth} for stratification under author citation counts. We also tabulate the thresholds corresponding to each quantile in these tables. 
We attempt to find reasonable thresholds for binning the strata while ensuring a non-zero sample size in each of them,
and covariates related to the subgroup stratification are discarded (\texttt{institute\_rank} and \texttt{author\_citations} respectively).
We acknowledge that some strata have few samples, making it hard for statistical procedures to be sensible, which renders a considerably wider confidence interval for estimated effect on the strata.

\section{Causal Inference: More Technical Details}\label{app:ci}

\subsection{Relaxing the DiD Assumption: Quantile-Quantile Equi-Confounding} \label{app:qqec}
In this section we explore an alternative to the
\emph{additive equi-confounding} assumption,
the core assumption in using NCO through DiD to deconfound the effect. Our motivation is to eliminate two undesirable consequences
that the additive equi-confounding assumption introduced:
\begin{enumerate}
    \item $N$ and $Y$ are assumed to be affected by the \emph{same} unobserved confounders; and
    \item unobserved confounders are assumed to affect $N$ and $Y$ at a similar scale. %
\end{enumerate}
In the analyses presented in the main text, we assume (1) and operationalize (2) by dichotomizing $N$ (thus $N$ is scaled to a similar level as $Y$).
As an alternative, \citet{Sofer2016OnNO} proposed the \emph{quantile-quantile equi-confounding} (QQ equi-confounding for short) assumption (for interested readers, this is an analogue in NCO of the more well-known change-in-change method studied in \citet{athey2006identification}), which is weaker than the additive equi-confounding assumption. QQ equi-confounding works when $Y$ and $N$ might be affected by a possibly different set of unobserved confounders, and is invariant to the scaling of $N$. In this sense, both (1) and (2) above can be relaxed in QQ equi-confounding.

We first collect some definitions and results from \citet{Sofer2016OnNO}.
Write $F_{X\mid Z}(x)$ as the cumulative distribution function (CDF) of $X$ given $Z$, and define $F^{-1}_{X\mid Z}$ as its inverse function. 
Assume for now we write $U$, $W$ for the unobserved confounder for $Y$ and $N$ respectively ($U$ and $W$ may also be associated with $A$ and between themselves),
the quantile-quantile (QQ) association between $U$ and $A$ conditional on $C$,
and between $W$ and $A$ conditional on $C$ are defined as:
\begin{align*}
    q_0(u\mid c) &= F_{U\mid A=0, C=c}\circ F^{-1}_{U\mid A=1, C=c}(u),\\
    q_1(w\mid c) &= F_{W\mid A=0, C=c}\circ F^{-1}_{W\mid A=1, C=c}(w),
\end{align*}
for $u,w\in[0,1]$.
The alternative QQ equi-confounding assumption is then framed as follows.
\begin{assumption}[QQ Equi-confounding]
\begin{align*}
q_0(v\mid c) = q_1(v\mid c), \quad v\in[0,1].
\end{align*}
\end{assumption}
This assumptions means, after transformed into the quantile level, the association between $U$ and $N$ is the same as between $W$ and $Y$,
conditional on $C$. Note that since quantiles are invariant under monotone transforms, this assumption does not restrict
the scale between $Y$ and $N$ to be the same.

Under the framework of QQ equi-confounding, an alternative positivity assumptions is needed. Let $N^*\sim (N\mid A=1, C)$ be a random variable distributed the same as the NCO in the treated group, positivity assumes that
if $0< f_{N\mid A=1,C}(N^*)$, then
$0< f_{N\mid A=0,C}(N^*)<1$, where $f$ is the probability density function corresponding to $F$. This assumption
is the analogue of positivity on the quantile level, which we assume.

\begin{figure}
    \centering
    \includegraphics[width=0.75\columnwidth]{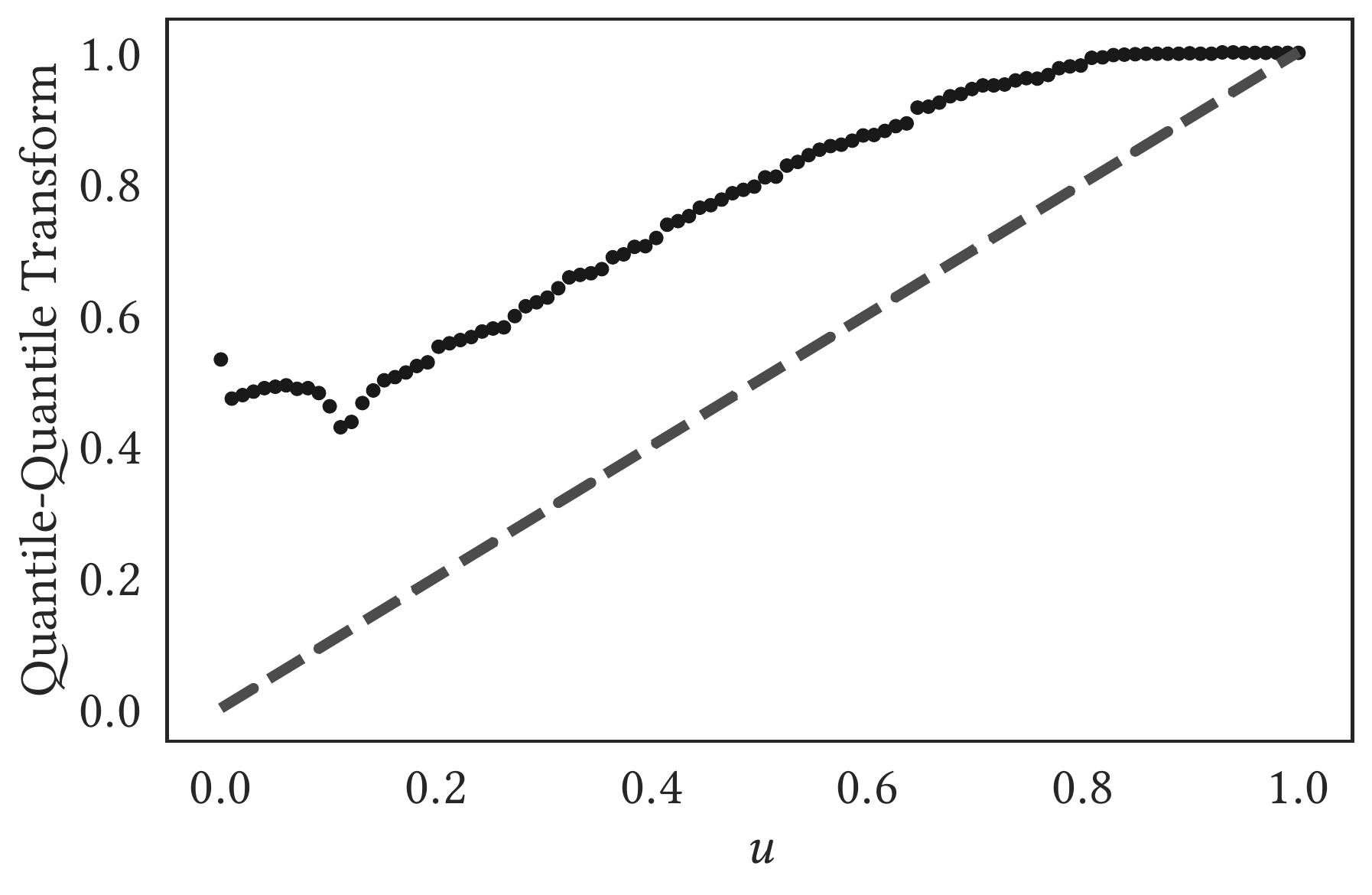}
    \caption{\textbf{Empirical QQ-Plot ($\operatorname{qq}(u)$).} The departure from the identity (dashed line) encodes unobserved confounding.}
    \label{fig:qqplot}
\end{figure}

\begin{theorem}[ATET Under QQ Equi-Confounding (Theorem 1 from \citealt{Sofer2016OnNO})]
Under the above assumptions, ATET can be expressed as:
\begin{align*}
\atet = \mathbb{E}[Y\mid A=1]-\mathbb{E}[\tilde{Y}],
\end{align*}
where
\begin{align*}
\tilde{Y}=F_{Y\mid A=0,C}^{-1} \circ F_{N\mid A=0,C}(N^*).
\end{align*}
\end{theorem}
As a first step, we examine the QQ transformation,
\[
\operatorname{qq}(u)=\mathbb{E}_C\left[\hat{F}_{Y\mid A=0,C}(\hat{F}^{-1}_{N\mid A=0,C}(u))\right]
\]
as a function of $u\in[0,1]$, where the expectation is taken with respect to covariates distributed as in the treated group.
We observe in \Cref{fig:qqplot} that the QQ plot departs from the identity (dashed) lines, which encodes unobserved confounding \citep{Sofer2016OnNO}.
Using $N=\cc^{(3)}$ as the NCO,
under the QQ equi-confounding assumption, 
we estimate that
\begin{align*}
    \atet = -4.375\%,
\end{align*}
with $95\%$-bootstrap confidence interval being
\begin{align*}
    (-9.965\%, -0.092\%).
\end{align*}
\paragraph{Discussion}
This confidence interval almost touches $0$, indicating that under the QQ equi-confounding assumptions,
estimated effects are very weak. Comparing the results with our analysis in \Cref{fig:fullsmp_fp},
though the direction of effect may differ, the significant levels are consistent:
in both analyses we do not have significant (or very weak) evidence for early arXiving to affect paper decision (either positively or negatively).

Although under the QQ equi-confounding setup,
we are able to circumvent the two limitations of the additive equi-confounding outlined at the beginning of this section, 
our choice of using the citation count $\cc^{(n)}$ might not be the best:
as discussed in \Cref{sec:intro}, the citation count itself
may not be the best NCO since early arXiving might as well
causally affects how it varies at a minor scale (e.g., to change it from a value of $10$ to $11$, though both may be view as ``less-cited''). A more elaborated study would entail ``smoothing out'' minor
variations in long-term citation counts while preserving as much information as it could. Such explorations are out of the scope of this appendix and we welcome future work along this direction.

\section{Dichotomize or Not?} \label{app:nco_discussion}
In our main analysis using DiD for NCO,
we dichotomize the citation counts $\cc^{(n)}$ at various thresholds $\cc^{(n)}_{q}$ defined by the empirical quantiles in the matched sample. There are two motivations for dichotomization.
\begin{enumerate}
    \item Dichotomization made $Y$ and $N$
    have the same scale, as required by the DiD additive equi-confounding assumption.
    \item As we view ``highly cited'' from long-term citation counts as an NCO, fluctuations at a minor scale (e.g., cited by $10$ times vs.~$11$ times) should be discarded. Dichotomization naturally ``smooth'' out these fluctuations.
\end{enumerate}
The limitation of dichotomization is also obvious: (1) we do not know what is the ``correct'' threshold for dichotomizing; (2) much information was lost during this process. To provide a comprehensive study, for each citation count $\cc^{(n)}$, we perform three studies using three NCOs by thresholding $\cc^{(n)}$ at three levels. The results are interpreted under the assumption of dichotomizing at the level is valid, which is uncheckable. On the other hand, to provide another perspective when all information is kept, in \Cref{app:qqec}, we explore an alternative methods in which $\cc^{(n)}$ is directly used as an NCO without dichotomizing.

\end{document}